# OPUS: An Efficient Admissible Algorithm for Unordered Search

**Geoffrey I. Webb**                                                    WEBB@DEAKIN.EDU.AU
*Deakin University, School of Computing and Mathematics*
*Geelong, Vic, 3217, Australia.*

## Abstract

OPUS is a branch and bound search algorithm that enables efficient admissible search through spaces for which the order of search operator application is not significant. The algorithm's search efficiency is demonstrated with respect to very large machine learning search spaces. The use of admissible search is of potential value to the machine learning community as it means that the exact learning biases to be employed for complex learning tasks can be precisely specified and manipulated. OPUS also has potential for application in other areas of artificial intelligence, notably, truth maintenance.

## 1. Introduction

Many artificial intelligence problems involve search. Consequently, the development of appropriate search algorithms is central to the advancement of the field. Due to the complexity of the search spaces involved, heuristic search is often employed. However, heuristic algorithms cannot guarantee that they will find the targets they seek. In contrast, an admissible search algorithm is one that is guaranteed to uncover the nominated target, if it exists (Nilsson, 1971). This greater utility is usually obtained at a significant computational cost.

This paper describes the OPUS (Optimized Pruning for Unordered Search) family of search algorithms. These algorithms provide efficient admissible search of search spaces in which the order of application of search operators is not significant. This search efficiency is achieved by the use of branch and bound techniques that employ domain specific pruning rules to provide a tightly focused traversal of the search space.

While these algorithms have wide applicability, both within and beyond the scope of artificial intelligence, this paper focuses on their application in classification learning. Of particular significance, it is demonstrated that the algorithms can efficiently process many common classification learning problems. This contrasts with the seemingly widespread assumption that the sizes of the search spaces involved in machine learning require the use of heuristic search.

The use of admissible search is of potential value in machine learning as it enables better experimental evaluation of alternative learning biases. Search is used in machine learning in an attempt to uncover classifiers that satisfy a learning bias. When heuristic search is used it is difficult to determine whether the search technique introduces additional implicit biases that cannot be properly identified. Such implicit biases may confound experimental results. In contrast, if admissible search is employed the experimenter can be assured that the search technique is not introducing confounding unidentified implicit biases into the experimental situation.

The use of OPUS for admissible search has already led to developments in machine learning that may not otherwise have been possible. In particular, Webb (1993) compared classifiers developed through true optimization of Laplace accuracy estimate with those obtained through heuristic search that sought but failed to optimize this measure. In general, the latter proved to have higher predictive accuracy than the former. This surprising result, that could not have been obtained without the use of admissible search, led Quinlan and Cameron-Jones (1995) to develop a theory of oversearching.

This paper offers two distinct contributions to the fields of computing, artificial intelligence and machine learning. First, it offers a new efficient admissible search algorithm for unordered search. Second, it demonstrates that admissible search is possible for a range of machine learning tasks that were previously thought susceptible only to efficient exploration through non-admissible heuristic search.

## 2. Unordered Search Spaces

For most search problems, the order in which operators are applied is significant. For example, when attempting to stack blocks it matters whether the red block is placed on the blue block before or after the blue block is placed on the green. When attempting to navigate from point A to point B, it is not possible to move from point C to point B before moving to point C. However, for some search problems, the order in which operators are applied is not significant. For example, when searching through a space of logical expressions, the effect of conjoining expression A with expression B and then conjoining the result with expression C is identical to the result obtained by conjoining A with C followed by B. Both sequences of operations result in expressions with equivalent meaning. In general, a search space is unordered if for any sequence $O$ of operator applications and any state $S$, all states that can be reached from $S$ by a permutation of $O$ are identical. It is this type of search problem, search through unordered search spaces, that is the subject of this investigation.

Special cases of search through unordered search spaces are provided by the subset selection (Narendra & Fukunaga, 1977) and minimum test-set (Moret & Shapiro, 1985) search problems. Subset selection involves the selection of a subset of objects that maximizes an evaluation criterion. The minimum test-set problem involves the selection of a set of tests that maximizes an evaluation criterion. Such search problems are encountered in many domains including machine learning, truth maintenance and pattern recognition. Rymon (1992) has demonstrated that Reiter's (1987) and de Kleer, Mackworth, and Reiter's (1990) approaches to diagnosis can be recast as subset selection problems.

The OPUS algorithms traverse the search space using a search tree. The root of the search tree is an initial state. Branches denote the application of search operators and the nodes that they lead to denote the states that result from the application of those operators. Different variants of OPUS are suited to each of optimization search and satisficing search. For optimization search, a goal state is an optimal solution. For satisficing search, a goal state is an acceptable solution. It is possible that a search space may include multiple goal states.

The OPUS algorithms take advantage of the properties of unordered search spaces to optimize the effect of any pruning of the search tree that may occur. In particular, when expanding a node $n$ in a search tree, the OPUS algorithms seek to identify search operators that can be excluded from consideration in the search tree descending from $n$ without excluding a sole goal node from that search tree. The OPUS algorithms differ from most previous admissible search algorithms employed in machine learning (Clearwater & Provost, 1990; Murphy & Pazzani, 1994; Rymon, 1992; Segal & Etzioni, 1994; Webb, 1990) in that when such operators are identified, they are removed from consideration in all branches of the search tree that descend from the current node. In contrast, the other algorithms only remove a single branch at a time without altering the operators considered below sibling branches, thereby pruning fewer nodes from the search space.

If it is not possible to apply an operator more than once on a path through the search space, search with unordered operators can be considered to be a subset selection problem—select a subset of operators whose application (in any order) leads to a goal state. If a single operator may be applied multiple times on a single path through the search space, search with unordered operators can be considered as a sub-multiset selection problem—select the multiset of operators whose application leads to the desired result.

A search tree that traverses an unordered search space in which multiple applications of a single operator are not allowed may be envisioned as in Figure 1. This example includes four search operators, named $a$, $b$, $c$ and $d$. Each node in the search tree is labeled by the set of operators by which it is reached. Thus, the initial state is labeled with the empty set. At depth one are all sets containing a single operator, at depth two all sets containing two operators and so on, up to depth four. Any two nodes with identical labels represent equivalent states.

There is considerable duplication of nodes in this search tree (the label $\{a, b, c, d\}$ occurs 24 times). In Figure 1 (and the following figures), the number of unique nodes is listed below each depth of the search tree. Where this number can be derived from the number of combinations to be considered, this derivation is also indicated.

It is common during search to prune regions of the search tree on the basis of investigations that determine that a goal state cannot lie within those regions. Figure 2 shows a search tree with the sub-tree below $\{c\}$ pruned. Note that, due to the duplication inherent in such a search tree, the number of unique nodes remaining in the tree is identical to that in the unpruned tree. However, if it has been deemed that no node descending from $\{c\}$ may be a goal, then all nodes elsewhere in the search tree that have identical labels (are reached via identical sets of operator applications) to any nodes that occur in the pruned region of the tree could also be pruned. Figure 3 shows the search tree remaining when all nodes below $\{c\}$ and all their duplicates have been deleted. It can be seen that the number of unique nodes in the remaining search tree (the tree at depths 2, 3 and 4) has been pruned by more than half. Similar results are obtained in the case where multiple applications of a single operator are allowed and the nodes are consequently labeled with multisets.

The OPUS algorithms do not provide pruning rules—mechanisms for identifying sections of the search tree that may be pruned. Rather, they take pruning rules as input and seek to optimize the effect of each pruning action that results from application of those rules.

The OPUS algorithms were designed for use with admissible pruning rules. When used solely with admissible pruning rules the algorithms are admissible. That is, they are guaranteed to find a goal state if one exists in the search space. However, the algorithms may also be used with non-admissible pruning heuristics to obtain efficient non-admissible search.

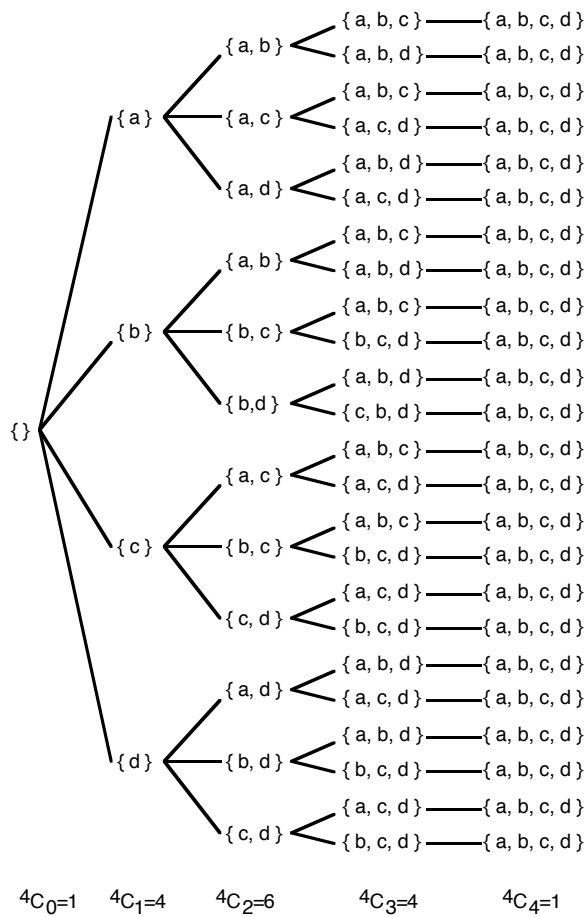

Figure 1: Simple unordered operator search tree

The OPUS algorithms are not only admissible (when used with admissible pruning rules), they are systematic (Pearl, 1984). That is, in addition to guaranteeing that a goal will be found if one exists, the algorithms guarantee that no state will be visited more than once during a search (so long as it is not possible to reach a single node by application of different sets of operators).

## 3. Fixed-order Search

A number of recent machine learning algorithms have performed restricted admissible search (Clearwater & Provost, 1990; Rymon, 1993; Schlimmer, 1993; Segal & Etzioni, 1994; Webb, 1990). All of these algorithms are based on an organization of the search tree, that, when considering the search problem illustrated in Figures 1 to 3, traverse the search space in the manner depicted in Figure 4. Such an organization is achieved by arranging the operators in a predefined order, and then applying at a node all and only operators that have a higher order than any operator that appears in the path leading to the node. This strategy will be called fixed-order search. (Fixed-order search has also been used for non-admissible search, for example, Buchanan, Feigenbaum, & Lederberg, 1971).

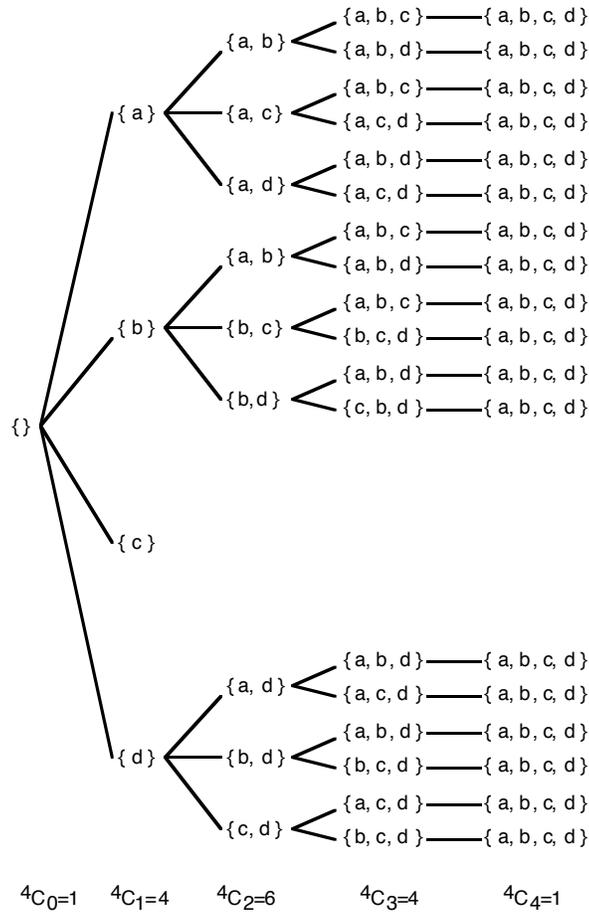

Figure 2: Simple unordered operator search tree with pruning beyond application of operator $c$

Figure 5 illustrates the effect of pruning the sub-tree descending below operator $c$, under fixed-order search. As can be seen, this is substantially less effective than the optimized pruning illustrated in Figure 3. Schlimmer (1993) ensures that the pruning effect illustrated in Figure 3 is obtained within the efficient search tree organization illustrated in Figure 4, by maintaining an explicit representation of all nodes that are pruned. The resulting search tree is depicted in Figure 6. This approach requires the considerable computational overhead of identifying and marking all pruned states following every pruning action, and the restrictive storage overhead of maintaining the representation. (One of the search problems tackled below contains $2^{162}$ states. To represent whether a state is pruned requires a single bit. Thus, $2^{162}$ bits would be required to represent the required information for this problem, a requirement well beyond the capacity of computational machinery into the foreseeable future.) Further, it is open to debate whether this approach does truly prune all identified nodes from the search space. Nodes that that have been 'pruned' will still need to be generated when encountered in previously unexplored regions of the search tree in order to be checked against the list of pruned nodes. Consider, for example, the node labeled {a} in Figure 5. When expanding this node it will be necessary to generate the node labeled {a, c}, even if this node has been marked as pruned. Only once it is generated is it possible to identify it as a node that has been 'pruned'. This node could in principle be pruned anyway by application of some variant of the technique that identified it as prunable in the first place. Viewed in this light, it can be argued that Schlimmer's (1993) approach does not reduce the number of nodes that must be generated under fixed-order search. All that it saves is the computational cost of determining for some nodes whether they require pruning or not. (This assumes that the optimistic pruning mechanism will be able to determine for any node $n$ from the search space below a pruned node $m$ that $n$ should also be pruned, irrespective of where $n$ is encountered in the search tree. If the optimistic pruning mechanism is deficient in that it cannot do this, then Schlimmer's (1993) approach will increase the amount of true pruning performed to the extent that it overcomes this deficiency.)

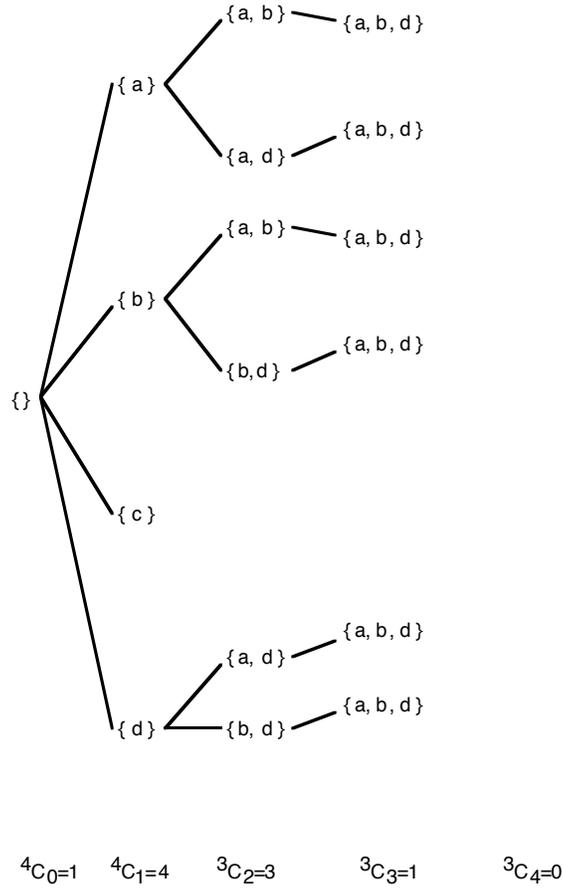

Figure 3: Simple unordered operator search tree with maximal pruning beyond application of operator $c$

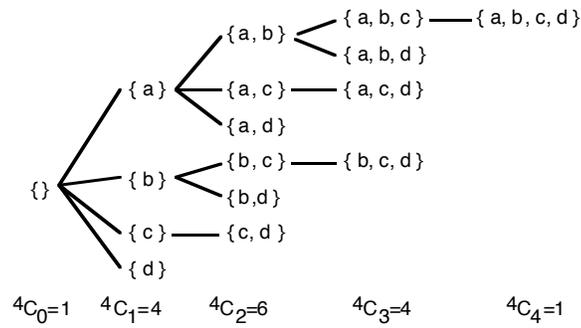

Figure 4: Static search tree organization used by fixed-order search

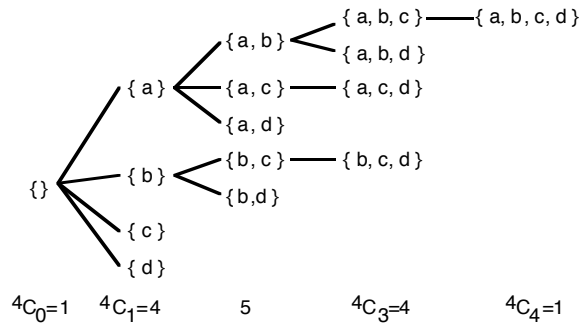

Figure 5: Effect of pruning under fixed-order search

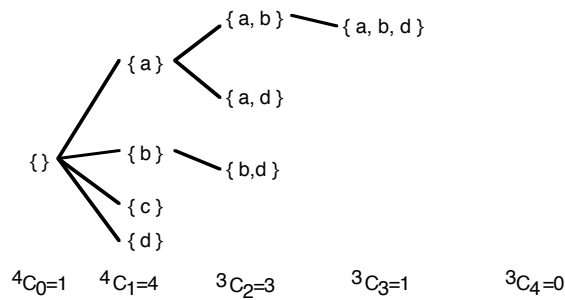

Figure 6: 'Optimal' pruning under fixed-order search

## 4. The Feature Subset Selection Algorithm

Fixed-order search traverses the search space in a naive manner—the topology of the search tree is determined in advance and takes no account of the efficiency of the resulting search. In contrast, the Feature Subset Selection (FSS) algorithm (Narendra & Fukunaga, 1977) performs branch and bound search in unordered search spaces, traversing the search space so as to visit each state at most once and dynamically organizing the search tree so as to maximize the proportion of the search space placed under unpromising operators. It can be viewed as a form of fixed-order search in which the order is altered at each node of the search tree so as to manipulate the topology of the search tree for the sake of search efficiency. Unlike Schlimmer (1993), the pruning mechanism ensures that nodes that are identified as prunable are not generated.

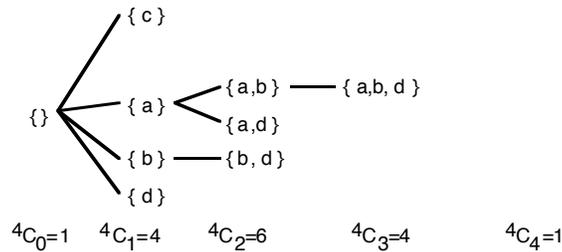

Figure 7: Pruning under FSS-like search

The power of this measure is illustrated by Figure 7. In this figure, fixed-order search is performed on the simple example problem illustrated in Figures 1 to 6, with the order changed so that the operator to be pruned, c, is placed first. As can be seen, this achieves the amount of pruning achieved by optimal pruning. This effect can be achieved with negligible computational or storage overhead.

However, FSS is severely limited in its applicability as—

- it is restricted to optimization search;

- it is restricted to tasks for which each operator may only be applied once (subset selection);

- it is restricted to search for a single solution;

- it requires that the values of states in the search space be monotonically decreasing. That is, the value of a state cannot increase as a result of an operator application; and

- the only form of pruning that it supports is optimistic pruning.

## 5. The OPUS Algorithms

The OPUS algorithms generalize the idea of search space reorganization from FSS. Two variations of OPUS are defined. OPUS[s] is a variant for satisficing search (search in which any qualified object is sought). OPUS[o] is a variant for optimization search (search in which an object that optimizes an evaluation function is sought). Whereas FSS uses node values for pruning, OPUS[o] uses optimistic evaluation of the search space below a node. This removes the requirement that the values of states in the search space be monotonically decreasing and opens the possibility of performing other types of pruning in addition to optimistic pruning.

In the analysis to follow, where comments apply equally to both variants the name OPUS will be employed. When a comment applies to only one variant of the algorithm, it will be distinguished by its respective superscript.

OPUS uses a branch and bound (Lawler & Wood, 1966) search strategy that traverses the search space in a manner similar to that illustrated in Figure 4 so as to guarantee that no two equivalent nodes in the search space are both visited. However, it organizes the search tree so as to optimize the effect of pruning, achieving the effect illustrated in Figure 6 without any significant computational or storage overhead.

Rather than maintaining an operator order, OPUS maintains at each node, $n$, the set of operators $n.active$ that can be applied in the search space below $n$. When the node is expanded, the operators in $n.active$ are examined to determine if any can be pruned. Any operators that can be pruned are removed from $n.active$. New nodes are then created for each of the operators remaining in $n.active$ and their sets of active operators are initialized so as to ensure that every combination of operators will be considered at only one node in the search tree.

It should be kept in mind that it is possible that many states in a search space may be goal states. For satisficing search all states that satisfy a given criteria are goal states. For optimization search, all states that optimize the evaluation criteria are goal states. For efficiency sake, the OPUS algorithms allow sections of the search space to be pruned even if they contain a goal state, so long as there remain other goal states in the remaining search space.

## 5.1 OPUS<sup>s</sup>

The OPUS[s] algorithm is presented in Figure 8. This description of OPUS[s] follows the conventions employed in the search algorithm descriptions provided by Pearl (1984).

This definition of OPUS[s] assumes that a single operator cannot be applied more than once along a single path through the search space. If an operator may be applied multiple times, the order of Steps 8a and 8b should be reversed. Unless otherwise specified, the following discussion of OPUS assumes that each operator may be applied at most once along a single path.

If it is desired to obtain all solutions that satisfy the search criterion,

- Step 2 should be altered to exit successfully, returning the set of all solutions;

- Step 6b should be altered to not exit, but rather to add the current node to the set of solutions; and

- The domain specific pruning mechanisms employed at Step 7 should also be modified so that no goal state may be pruned from the search space.

This form of search could be used in an assumption-based truth maintenance system to find the set of all maximally general consistent assumptions. This would provide efficient search without the need to maintain and search an explicit database of inconsistent assumptions such as the ATMS no-good set (de Kleer et al., 1990). Unless otherwise specified, the discussion of OPUS below assumes that a single solution is sought.

The algorithm does not specify the order in which nodes should be selected for expansion at Step 3. Nodes may be selected at random, by a domain specific selection function, or by

**Data structure:**

Each node, $n$, in the search tree has associated with it three items of information:

$n.state$  the state from the search space that is associated with the node;

$n.active$  the set of operators to be explored in the sub-tree descending from the node; and

$n.mostRecentOperator$  the operator that was applied to the parent node's state to create the current node's state.

**Algorithm:**

1. Initialize a list called $OPEN$ of unexpanded nodes as follows,

    (a) Set $OPEN$ to contain one node, the start node $s$.

    (b) Set $s.active$ to the set of all operators, $\{o_1, o_2, ...o_n\}$

    (c) Set $s.state$ to the start state.

2. If $OPEN$ is empty, exit with failure; no goal state exists.

3. Remove from $OPEN$ a node $n$, the next node to be expanded.

4. Initialize to $n.active$ a set containing those operators that have yet to be examined, called $RemainingOperators$.

5. Initialize to $\{\}$ a set of nodes, called $NewNodes$, that will contain the descendants of $n$ that are not pruned.

6. Generate the children of $n$ by performing the following steps for every operator $o$ in $n.active$,

    (a) Generate $n'$, a node for which $n'.state$ is set to the state formed by application of $o$ to $n.state$.

    (b) If $n'.state$ is a goal state, exit successfully with the goal represented by $n'$.

    (c) Set $n'.mostRecentOperator$ to $o$.

    (d) Add $n'$ to $NewNodes$.

7. While there is a node $n'$ in $NewNodes$ such that pruning rules determine that no sole remaining goal state is accessible from $n'$ using only operators in $RemainingOperators$, prune all nodes in the search tree below $n'$ from the search tree below $n$ as follows,

    (a) Remove $n'$ from $NewNodes$.

    (b) Remove $n'.mostRecentOperator$ from $RemainingOperators$.

8. Allocate the remaining operators to the remaining nodes by processing each node $n'$ in $NewNodes$ in turn as follows,

    (a) Remove $n'.mostRecentOperator$ from $RemainingOperators$.

    (b) Set $n'.active$ to $RemainingOperators$.

9. Add the nodes in $NewNodes$ to $OPEN$.

10. Go to Step 2.

Figure 8: The OPUS[s] Algorithm

the order in which nodes are placed in $OPEN$. First-in-first-out node selection results in breadth-first search while last-in-first-out node selection results in depth-first search.

The order of processing is also unspecified at Steps 7, 8 and 9. Depending upon the domain, practical advantage may be obtained by specific orderings at these steps.

OPUS$^s$ has been used in a machine learning context to search the space of all generalizations that may be formed through deletion of conjuncts from a highly specific classification rule. The goal of this search is to uncover the set of all most general rules that cover identical objects in the training data to those covered by the original rule (Webb, 1994a).

### 5.2 OPUS$^o$

A number of changes are warranted if OPUS is to be applied to optimization search. The following definition of OPUS$^o$, a variant of OPUS for optimization search, assumes that two domain specific functions are available. The first of these functions, $value(n)$, returns the value of the state for node $n$, such that the higher the value returned, the higher the preference for the state[1]. The second function, $optimisticValue(n, o)$ returns a value such that if there exists a node, $b$, that can be created by application of any combination of operators in the set of operators $o$ to the state for node $n$, and $b$ represents a best solution (maximizes $value$ for the search space), $optimisticValue(n, o)$ will be no less than $value(b)$. This is used for pruning sections of the search tree. In general, the lower the values returned by $optimisticValue$, the greater the efficiency of pruning. At any time, it is possible to prune any node with an optimistic value that is less than or equal to the best value of a node explored to date.

OPUS$^o$ is able to take advantage of the presence of optimistic values to further optimize the effect of pruning beyond that obtained solely by maximizing the proportion of the search space placed under nodes that are immediately pruned. Generalizing a heuristic used in FSS, nodes with lower optimistic values are given more active operators and thus have greater proportions of the search space placed beneath them than nodes with higher optimistic values. This is achieved by the order of processing at Step 9. The rationale for this strategy is that the lower the optimistic value the higher the probability that the node and its associated search tree will be pruned before it is expanded. Maximizing the proportion of the search space located below nodes with low optimistic value maximizes the proportion of the search space to be pruned and thus not explicitly explored.

Figure 9 illustrates this effect with respect to a simple machine learning task—search for a propositional expression that describes the most target examples and no non-target examples. The seven search operators each represent conjunction with a specific proposition $male$, $female$, $single$, $married$, $young$, $mid$ and $old$, respectively. Search starts from the expression $anything$. A total of 128 expressions may be formed by conjunction of any combination of these expressions. Twelve objects are defined:

<div align="center">

*male, single, young, TARGET*

*male, single, mid, TARGET*

*male, single, old, TARGET*

*male, married, young, NON-TARGET*

</div>

---

[1]. For ease of exposition, it will be assumed that optimization means maximization of a value. It would be trivial to transform the algorithm and discussion to accommodate other forms of optimization.

*male, married, mid, NON-TARGET*
*male, married, old, NON-TARGET*
*female, single, young, NON-TARGET*
*female, single, mid, NON-TARGET*
*female, single, old, NON-TARGET*
*female, married, young, NON-TARGET*
*female, married, mid, NON-TARGET*
*female, married, old, NON-TARGET.*

Of these objects, the first three are distinguished as targets. The value of an expression is determined by two functions, *negCover* and *posCover*. The *negCover* of an expression is the number of non-target objects that it matches. The *posCover* of an expression is the number of target objects that it matches. The expression *anything* matches all objects. The value of an expression is $-\infty$ if *negCover* is not equal to zero. Otherwise the value equals *posCover*. This preference function avoids expressions that cover any negative objects and favors, of those expressions that cover no negative objects, those expressions that cover the most positive objects. The optimistic value of a node equals the *posCover* of the node's expression.

Figure 9 depicts the nine nodes considered by OPUS$^{\mathrm{O}}$ for this search task. For each node the following are listed:

- the expression;

- the number of target and the number of non target objects matched (cover);

- the value;

- the potential value; and

- the operators placed in the node's set of active operators and hence included in the search tree below the node.

The search space is traversed as follows. The first node, *anything*, is expanded, producing its seven children for which values and optimistic values are determined. No node can be pruned as all have potential values greater than the best value so far encountered. The active operators are then distributed, maximizing the proportion of the search space placed below nodes with low optimistic values. Of the two nodes with the highest optimistic values, *male* and *single*, one receives no active operator and the other receives the first as its sole active operator. One or the other is then expanded. If it is the one with no active operators, *single*, no further nodes are generated. Then the other, *male*, is expanded, generating a single node, *male* $\wedge$ *single*, with a value of 3. Immediately this node is generated, all remaining open nodes can be pruned as none has an optimistic value greater than this new maximum value, 3.

Note that no nodes can be pruned until the node for *male* $\wedge$ *single* is considered as, up to that point, no node has been encountered with a lower optimistic value than the best actual value. Consequently, if the search tree was not distributed in accord with potential value, the set of active operators for the node male would be $\{female, single, married, young, mid, old\}$. Instead of considering a single node when *male* was expanded, it would be necessary to consider six nodes. If the search space was more complex and continued to depth three or beyond, there would be a commensurate increase in the proportion of the search space explored unnecessarily.

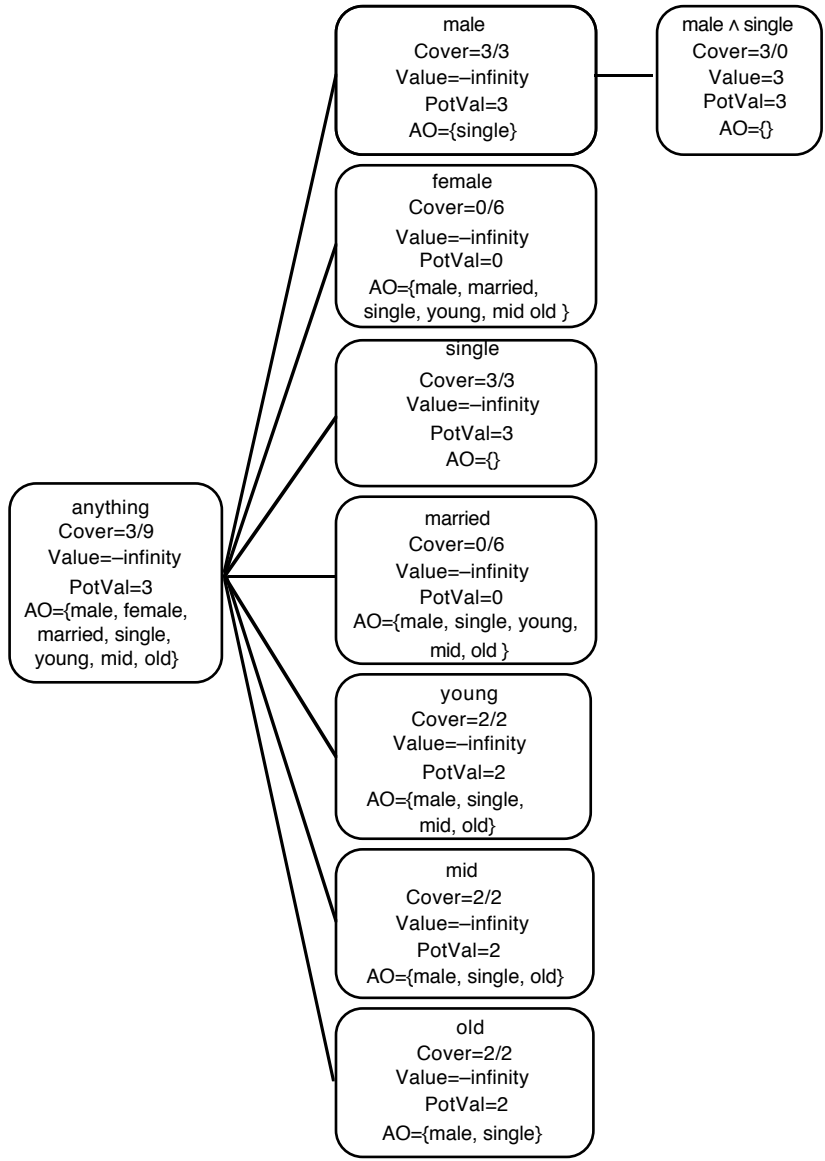

Figure 9: Effect of pruning when search tree ordered on optimistic value

Note also that the search in this example does not terminate when the goal node is first encountered, as the system cannot determine that it is a goal node until all other nodes that might have higher values have been explored or pruned.

### 5.2.1 THE OPUS$^O$ ALGORITHM

OPUS$^O$, the algorithm for achieving the above effect, can be defined as in Figure 10. Note that optimistic pruning need not be performed at Step 8 as it is performed at Step 10a, irrespective.

This definition of OPUS$^O$ assumes that a single operator cannot be applied more than once along a single path through the search space. To allow multiple applications of a single operator, the order of Steps 9a and 9b should be reversed.

The algorithm could also be modified to identify and return all maximal solutions through a modification similar to that outlined above to allow OPUS$^S$ to return all solutions.

It is possible to further improve the performance of OPUS$^O$ if there is a lower limit on an acceptable solution. Then, the objective of the search is to find a highest valued node so long as that value is greater than a pre-specified minimum. In this case, all nodes whose potential value is less than or equal to the minimum may also be pruned at Step 10a.

Like OPUS$^S$, OPUS$^O$ does not specify the order in which nodes in OPEN should be expanded (Step 4). Selection of a node with the highest optimistic value will minimize the size of the search tree. If there is a single node $n$ that optimizes the optimistic value, the search cannot terminate until $n$ has been expanded. This is because no node with a lower optimistic value may yield a solution with a value higher than the optimistic value of $n$. However, an expansion of $n$ may yield a solution that has a value higher than other candidate's optimistic values, allowing those other candidates to be discarded without expansion. Thus, selecting a single node with the highest optimistic value is optimal with respect to the number of nodes expanded because it maximizes the number of nodes that may be pruned without expansion. Where multiple nodes all maximize the optimistic value, at least one of these must be expanded before the search can terminate (and then the search will only terminate if expansion of that node leads to a node with a value equal to that optimistic value.)

In many cases it is more important to consider the number of nodes explored by an algorithm, rather than the number of nodes expanded. A node is explored if it is evaluated. Every time a node is expanded, all of its children will be explored. Many of these children may be pruned, however, and never be expanded. In addition to minimizing the number of nodes expanded, this form of best-first search will also minimize the number of nodes explored (within the constraint that where nodes have equal optimistic values it is not possible to anticipate which one to select in order to minimize the number of nodes explored). This is due to the strategy that the algorithm employs to distribute operators beneath nodes. The nodes that OPUS$^o$ expands under best-first search will be those with highest optimistic value. OPUS$^O$ always allocates fewer active operators to a node with higher optimistic value than to a node with lower optimistic value. The number of nodes examined when a node

**Algorithm:**

1. Initialize a list called $OPEN$ of unexpanded nodes as follows,

   (a) Set $OPEN$ to contain one node, the start node $s$.

   (b) Set $s.active$ to the set of all operators, $\{o_1, o_2, ...o_n\}$

   (c) Set $s.state$ to the start state.

2. Initialize $BEST$, the best node examined so far, to $s$.

3. If $OPEN$ is empty, exit successfully with the solution represented by $BEST$.

4. Remove from $OPEN$ a node $n$, the next node to be expanded.

5. Initialize to $n.active$ a set containing those operators that have yet to be examined, called $RemainingOperators$.

6. Initialize to $\{\}$ a set of nodes, called $NewNodes$, that will contain the descendants of $n$ that are not pruned.

7. Generate the children of $n$ by performing the following steps for every operator $o$ in $n.active$,

   (a) Generate $n'$, a node for which $n'.state$ is set to the state formed by application of $o$ to $n.state$.

   (b) If $value(n')$ is greater than $value(BEST)$
       i. Set $BEST$ to $n'$.
       ii. Prune the search tree by removing from $OPEN$ all nodes $x$ such that $optimisticValue(x, x.active)$ is less than $value(BEST)$.

   (c) Add $n'$ to $NewNodes$.

   (d) Set $n'.mostRecentOperator$ to $o$.

8. While there is a node $n'$ in $NewNodes$ such that pruning rules determine that no sole remaining goal state is accessible from $n'$ using only operators in $RemainingOperators$, prune all nodes in the search tree below $n'$ from the search tree below $n$ as follows,

   (a) Remove $n'$ from $NewNodes$.

   (b) Remove $n'.mostRecentOperator$ from $RemainingOperators$.

9. Allocate the remaining operators to the remaining nodes by processing each node $n'$ in $NewNodes$ in turn as follows, each time selecting the previously unselected node that minimizes $optimisticValue(n', RemainingOperators)$,

   (a) Remove $n'.mostRecentOperator$ from $RemainingOperators$.

   (b) Set $n'.active$ to $RemainingOperators$.

10. Perform optimistic pruning while adding the remaining nodes to $OPEN$ by processing each node $n'$ in $NewNodes$ in turn as follows,

   (a) If $optimisticValue(n', n'.active)$ is greater than $value(BEST)$,
       i. Add $n'$ to $OPEN$.

11. Go to Step 3.

Figure 10: The OPUS$^\text{o}$ algorithm

$n$ is expanded equals the number of active operators at $n$. Hence, the number of nodes examined for those nodes expanded will be minimized (within the constraints of the use only of information that can be derived from the current state and the operators that are active at that state).

However, while this best-first approach minimizes the number of nodes expanded, it may not be storage optimal due to the large potential storage overheads. If the storage overhead is of concern, depth-first rather than best-first traversal may be employed, at the cost of a potential increase in the number of nodes that must be expanded. If depth-first search is employed, nodes should be added to $OPEN$ by order of optimistic value at Step 10. This will ensure that nodes open at a single depth will be expanded in a best-first manner, with the benefits outlined above.

### 5.2.2 Relation to Previous Search Algorithms

OPUS$^O$ can be viewed as an amalgamation of FSS (Narendra & Fukunaga, 1977) with A* (Hart, Nilsson, & Raphael, 1968). FSS performs branch and bound search in unordered search spaces, traversing the search space so as to visit each state at most once and dynamically organizing the search tree so as to maximize the proportion of the search space placed under unpromising operators. However, FSS requires that the values of states in the search space be monotonically decreasing. That is, the value of a state cannot increase as a result of an operator application. OPUS$^O$ generalizes from FSS by employing both the actual values of states and optimistic evaluation of nodes in the search tree, in a manner similar to A*. In consequence, there are only two minor constraints upon the values of states and the optimistic values of nodes in the search spaces that OPUS$^O$ can search. These are the requirements that—

- for at least one goal state $g$ and for any node $n$, if $g$ lies below a node $n$ in the search tree, the optimistic value of $n$ be no lower than the value of $g$; and

- that any and only states of maximal value be goal states.

It follows that OPUS$^O$ has wider applicability than FSS.

OPUS$^O$ also differs from FSS by integrating pruning mechanisms other than optimistic pruning into the search process. This facility is crucial when searching large search spaces such as those encountered in machine learning.

A further innovation of the OPUS algorithms is the use of the restricted set of operators available at a node in the search tree to enable more focused pruning than would otherwise be the case. There may be circumstances in which it would be possible to reach a goal from the state at a node $n$, but only through application of operators that are not active at $n$. The pruning rules are able to take account of the active operators to provide pruning in this context—pruning that would not otherwise be possible. Similarly, the set of active operators can be used to calculate a more concise estimate of the optimistic value than would otherwise be possible.

OPUS$^O$ differs from A* in the manner in which it dynamically organizes the search tree so as to maximize the proportion of the search space placed under unpromising operators. It also differs from A* in that A* relies upon the value of a node being equivalent to the sum of costs of the operations that lead to that node, whereas OPUS allows any method for determining a node's value.

Rymon (1993) discusses dynamic organization of the search tree during admissible search through unordered search spaces for the purpose of altering the topology of the data structure (SE-tree) produced. This contrasts with the use of dynamic organization of the search tree in OPUS[o] to increase search efficiency.

## 5.3 OPUS and Non-admissible Search

As was pointed out above, although the OPUS algorithms were designed for admissible search, if they are applied with non-admissible pruning rules they may also be used for non-admissible search. This may be useful if efficient heuristic search is required. Most non-admissible heuristic search strategies embed the heuristics in the search technique. For example, beam search relies upon the use of a fixed maximum number of alternative options that are to be considered at any stage during the search. The heuristic is to prune all but the $n$ best solutions at each stage during search. The precise implications of this heuristic for a particular search task may be difficult to evaluate. In contrast, the use of OPUS with non-admissible pruning rules places the non-admissible heuristic in a clearly defined rule which may be manipulated to suit the circumstances of a particular search problem.

Another feature of OPUS[o] is that at all stages it has available the best solution encountered to date during the search. This means that the search can be terminated at any time. When terminated prematurely, the current best solution would be returned on the understanding that this solution may not be optimal. If the algorithm is to be employed in this context, it may be desirable to employ best-first search, opening nodes with highest actual (as opposed to optimistic) value first, on the assumption that this should lead to early investigation of high valued nodes.

## 5.4 Complexity and Efficiency Considerations

OPUS ensures that no state is examined more than once (unless identical states can be formed by different combinations of operator applications), using a similar search tree organization strategy to that of fixed-order search. It differs, however, in that instead of placing the largest subsection of the search space under the highest ordered operator, the second largest subsection under the second highest ordered operator, and so on, whenever pruning occurs, the largest possible proportion of the search space is placed under the pruned node, and hence is immediately pruned.

If there are $n$ operators active at the node $e$ being expanded, the search tree below and including that node will contain every combination of any number of those operators (the application of none of the operators results in $e$). Thus, the search tree below and including $e$ will contain $2^n$ nodes. Exactly half of these, $2^{n-1}$, will have a label including any single operator $o$. OPUS ensures that if any operator $o$ is pruned when a node $e$ is expanded, that all nodes containing $o$ are removed from the search tree below $e$ and are never examined (except, of course, the node reached by a single application of $o$ that must be examined in order to determine that $o$ should be pruned). Thus, the search tree below a node is almost exactly halved if a single operator can be pruned. Each subsequent operator pruned at that node reduces the remaining search tree by the same proportion. Thus, the size of the remaining search tree is divided by almost exactly $2^p$, where $p$ is the number of operators pruned.

In contrast, the number of nodes pruned under fixed-order search depends upon the ranking of the operator within the fixed operator ranking scheme. Only for the highest ranked operator will the same proportion of the search tree be pruned as under OPUS. In general, when an operator is pruned, only those nodes whose labels include that operator in combination exclusively with lower ranked operators will be pruned. This effect is illustrated in Figure 5 in which pruning below $\{c\}$ removes only $\{c, d\}$ from the search tree. Thus, $2^{n-r} - 1$, nodes are immediately pruned from the search tree, where $n$ is the number of operators active at the node being expanded and $r$ is the ranking within those operators of the operator being pruned, with the highest rank being 1. This contrasts with the $2^{n-1} - 1$ nodes pruned by OPUS.

However, the difference in the number of nodes explored under the two strategies is not quite as great as this analysis might suggest, as (assuming the availability of a reasonable optimistic pruning mechanism) fixed-order search can also prune the operator every time that it is examined deeper in the search tree in combination with higher ranked operators. Thus, in Figure 5, when they were eventually examined, pruning would occur at nodes $\{a, b, c\}$, $\{a, c\}$ and $\{b, c\}$. Thus, $\{a, b, c, d\}$, $\{a, c, d\}$ and $\{b, c, d\}$ would also be pruned from the search tree. In other words, under fixed-order search, if an operator is pruned it will not be considered in combination with any lower ranked operator, but will be considered with every combination of any number of higher ranked operators. There are $2^{r-1}$ combinations of higher ranked operator. It follows that fixed-order search considers this many more nodes than OPUS when a single operator is pruned. Thus, for each operator that can be pruned at a node $n$, OPUS explores $2^{r-1}$ less nodes below $n$ than fixed-order search.

As the rank order of the operators pruned will tend to grow as the number of operators grows, it follows that, in the average case, the advantage accrued from the use of OPUS will grow exponentially as the number of operators grows. OPUS will tend to have the greatest relative advantage for the largest search spaces.

Note that OPUS is not always able to guarantee that the maximal possible pruning occurs as the result of a single pruning action. For example, if OPUS is being used to search the space of subsets of a set of items, and it can be determined that no superset of the set $s$ at a node may be a solution, but some items are not active at the current node, supersets of $s$ that contain the items that are not active may be explored elsewhere in the search tree. An algorithm that could prune all such supersets could perform more pruning than OPUS. While it might be claimed that Schlimmer's (1993) search method performs such pruning, it should be recalled that it does not prevent the 'pruned' nodes from being generated elsewhere in the tree, but rather, ensures that such nodes are pruned once generated. OPUS, if armed with suitable pruning rules, should also be able to prune such nodes when encountered. OPUS maximizes the pruning performed within the constraints of the localized information to which it has access.

However, while the constraint provided by the active operators prevents OPUS from performing some pruning, it also enables it to perform other pruning that would not otherwise be possible. This is because it is only necessary when considering whether to prune a node to determine whether nodes that can be reached by active operators may contain a solution. Thus, to continue the example of subset search, even when supersets of the set at the current node $n$ are potential solutions, it will still be possible to prune the search tree below $n$ if all of the supersets that are potential solutions contain items that are not active at $n$. Schlimmer's (1993) approach does not allow pruning in such a context.

To illustrate this effect, let us revisit the search space examined in Figures 1 to 7. Even though the search space below $\{c\}$ has been pruned, 'optimal pruning' cannot take this into account in its optimistic evaluations of other nodes as there is no mechanism by which this information can be communicated to the optimistic evaluation function (other than by actually exploring the space below the node to be evaluated, which defeats the purpose of optimistic evaluation). For example, when evaluating the optimistic value of the node $\{a\}$, the optimistic evaluation function cannot return a different value than would be the case if $\{c\}$ had not been pruned. By contrast, the optimistic evaluation function employed by OPUS[o] can take account of this by taking the active operators for the current node into consideration. Such an optimistic evaluation function is described in Section 6.1 below. It will often be possible to use the information that particular operators are not available in the search tree below a node to substantially improve the quality of the optimistic evaluation of that node.

It should also be noted that no algorithm that does not employ backtracking can guarantee that it will minimize the number of nodes expanded under depth-first search. If a poor node is chosen for expansion under depth-first search, the system is stuck with having to explore the search space below that node before it can return to explore alternatives. No algorithm can guarantee against a poor selection unless the optimistic evaluation function has high enough accuracy to prevent the need for backtracking. It follows that no algorithm that requires backtracking can guarantee that it will minimize the number of nodes that are expanded. Thus, OPUS is heuristic with respect to minimizing computational complexity under depth-first search.

The storage requirements of OPUS will depend upon whether depth, breadth or best-first search is employed. If depth-first search is employed, the maximum storage requirement will be less than the maximum depth of the search tree multiplied by the maximum branching factor. However, if breadth or best-first search is employed, in the worst case, the storage requirement is exponential. At any stage during the search, the storage requirement is that of storing the frontier nodes of the search. The number of frontier nodes cannot exceed the number of leaf nodes in the complete search tree. For search in which no operator may be applied more than once (subset selection), if there is no pruning, the number of leaf nodes is $2^{n-1}$, where $n$ is the number of operators. This assertion can be justified as follows. If the order in which operators are considered is invariant, all nodes reached via the last operator considered will be a leaf node. As the search is admissible, the last operator must be considered with every combination of other operators. There are $2^{n-1}$, other combinations of operators. The order in which operators are considered will not alter the number of leaf nodes in the absence of pruning. For search in which there is no limit on the number of applications of a single operator (sub-multiset selection) there is no upper limit on the potential storage requirements.

Irrespective of the storage requirements, in the worst case OPUS will have to explore every node in the search space. This will only occur if no pruning is possible during a search. If operators can only be applied once per solution, the number of nodes in the search space will equal $2^n$, where $n$ is the number of operators. Thus, the worst case computational complexity of OPUS is exponential, irrespective of whether depth, breadth or best-first search is employed.

OPUS is clearly inappropriate, both in terms of computational and, when using breadth or best-first search, storage requirements, for search problems in which substantial proportions of the search space cannot be pruned. For domains in which substantial pruning is possible, however, the average case complexity (computational and/or storage) may turn out to be polynomial. Experimental evidence that this is indeed the case for some machine learning tasks is presented below in Section 6.3.

### 5.5 How the Search Efficiency of OPUS Might be Improved

As is noted in Section 5.4, the OPUS algorithms are not always able to guarantee that the maximum possible amount of pruning is performed. As noted, one restriction upon the amount of pruning performed is the localization inherent in the use of active operators. While this localization allows some pruning that would not otherwise be possible, it also has the potential to restrict the number of supersets of the set of operators at a pruned node that are also pruned. There may be value in developing mechanisms that enable such pruning to be propagated beyond the node at which a pruning action occurs and the sub-tree below that node.

Another aspect of the algorithms that has both positive and negative aspects is the type of information returned by the pruning mechanisms. These mechanisms allow the pruning of any branch of the search tree so long as at least one goal is not below that branch. This contrasts with an alternative strategy of only pruning branches that do not lead to any goal. The strategy used can be beneficial, as it maximizes the amount of pruning that can be performed. However, it is always possible that a branch containing a goal that could be found with little exploration will be pruned in favor of a branch containing a goal that requires extensive exploration to uncover. There is potential for gain through augmenting the current pruning mechanisms with means of estimating the search cost of uncovering a goal beneath each branch in a tree.

## 6. Evaluating the Effectiveness of the OPUS Algorithms

Theoretical analysis has demonstrated that OPUS will explore fewer nodes than fixed-order search and that the magnitude of this advantage will increase as the size of the search space increases. However, the precise magnitude of this gain will depend upon the extent and distribution within the search tree of pruning actions. Of further interest, there are a number of distinct elements to each of the OPUS algorithms, including—optimistic pruning; other pruning (pruning in addition to optimistic pruning); dynamic reorganization of the search tree; and maximization of the proportion of the search space placed under nodes with low optimistic value. The following experiments evaluate the magnitude of the advantage to OPUS obtained for real world search tasks and explore the relative contribution of each of the distinct elements of the OPUS algorithms.

To this end, OPUS$^o$ was applied to a class of real search tasks—finding pure conjunctive expressions that maximize the Laplace accuracy estimate with respect to a training set of preclassified example objects. This is, for example, the search task that CN2 purports to heuristically approximate (Clark & Niblett, 1989) when forming the disjuncts of a disjunc-tive classifier. Machine learning systems have employed OPUS$^o$ in this manner to develop rules for inclusion both in sets of decision rules (Webb, 1993) and in decision lists (Webb, 1994b). (The current experiments were performed using the Cover learning system, which, by default, performs repeated search for pure conjunctive classifiers within a CN2-like covering algorithm that develops disjunctive rules. This more extended search for disjunctive rules was not used in the experiments, as it makes it difficult to compare alternative search algorithms. This is because, if two alternative algorithms find different pure conjunctive rules for the first disjunct, their subsequent search will explore different search spaces.)

Numerous efficient admissible search algorithms exist for developing classifiers that are consistent with a training set of examples. The two classic algorithms for this purpose are the least generalization algorithm (Plotkin, 1970) and the version space algorithm (Mitchell, 1977). The least generalization algorithm finds the most specialized class description that covers all objects in a training set containing only positive examples. The version space algorithm finds all class descriptions that are complete and consistent with respect to a training set of both positive and negative examples. Hirsh (1994) has generalized the version space algorithm to find all class descriptions that are complete and consistent to within defined bounds of the training examples. The least generalization and version space algorithms will usually require a strong inductive bias in the class description language (restriction on the types of class descriptions that will be considered) if they are to find useful class descriptions (Mitchell, 1980). SE-tree-based learning (Rymon, 1993) demonstrates admissible search for a set of consistent class descriptions within more complex class description languages than may usefully be employed with the least generalization or version space algorithms. Oblow (1992) describes an algorithm that employs admissible search for pure conjunctive terms within a heuristic outer search for k-DNF class descriptions that are consistent with the training set.

However, for many learning tasks it is desirable to consider class descriptions that are inconsistent with the training set. One reason for this is that the training set may contain noise (examples that are inaccurate). Another reason is that it may not be possible to accurately describe the target class in the available language for expressing class descriptions. In this case it is necessary to consider approximations to the target class. A further reason is that the training set may contain insufficient information to reliably determine the exact class description. In this case, the best solution may be an approximation that is known to be incorrect but for which there is strong evidence that the level of error is low.

Both Clearwater and Provost (1990) and Segal and Etzioni (1994) use admissible fixed-order search to explore classifiers that are inconsistent with the training set. However, the admissible search of Clearwater and Provost (1990) is not computationally feasible for large search spaces. Segal and Etzioni (1994) bound the depth of the search space considered in order to maintain computational tractability. Smyth and Goodman (1992) use optimistic pruning to search for optimal rules, but do not structure their search to ensure that states are not searched multiple times. No other previous admissible search algorithm has been employed in machine learning to find classifiers that are inconsistent with the training set and maximize an arbitrary preference function. The following experiments seek to demonstrate that such search is feasible using OPUS.

Where it is allowed that a class description may be inconsistent with the training set, it is helpful to employ an explicit preference function. Such a function is applied to a class description and returns a measure of its desirability. This evaluation will usually take account of how well the description fits that training set and may also include a bias toward particular types of class descriptions, for example, a preference for syntactic simplicity. Such a preference function expresses an inductive bias (Mitchell, 1980).

OPUS[o] may be employed for admissible search in such contexts, provided a search space can be defined that may be traversed by a finite number of unordered search operators. For example, OPUS[o] may be employed to search for a class description in a language of pure-conjunctive descriptions by examining a search space starting with the most general possible class description *true* and employing search operators, each of which has the effect of conjoining a specific clause to the current description. Such search may be performed with an arbitrary preference function, provided appropriate optimistic evaluation functions can be defined.

The next section describes experiments in which OPUS[o] was applied in this manner.

## 6.1 The Search Task

The pure conjunctive expressions consisted of conjunctions of clauses of the form *attribute $\neq$ value*. For attributes with more than two values, such a language is more expressive than a language allowing only conjunctions of clauses of the form *attribute = value*. Indeed, it has equivalent expressiveness to a language that supports internal disjunction. For example, with respect to an attribute $a$ with the values $x$, $y$ and $z$, a language restricted to conjunctions of equality expressions cannot express $a \neq x$, whereas a language restricted to conjunctions of inequality expressions can express $a = x$ using the expression $a \neq y \wedge a \neq z$. In internal disjunctive (Michalski, 1984) terms, $a \neq x$ is equivalent to $a = y$ or $z$.

It should be noted that—

- For attributes with more than two values the search space for conjunctions of inequality expressions is far larger than the search space for conjunctions of equality expressions. For each attribute, the size of the search space is multiplied by $2^n$ for the former and by $n + 1$ for the latter, where $n$ is the number of values for the attribute.

- The software employed in this experimentation can also be used to search the smaller search spaces of equality expressions with the same effects as are demonstrated in the following experiments.

Search starts from the most general expression, true. Each operator performs conjunction of the current expression with a term $A \neq v$, where $A$ is an attribute and $v$ is any single value for that attribute.

The Laplace (Clark & Boswell, 1991) preference function was used to determine the goal of the search. This function provides a conservative estimate of the predictive accuracy of a class description, $e$. It is defined as

$$value(e) = \frac{posCover(e) + 1}{posCover(e) + negCover(e) + noOfClasses}$$

where $posCover(e)$ is the number of positive objects covered by $e$; $negCover(e)$ is the number of negative objects covered by $e$; and $noOfClasses$ is the number of classes for the learning task.

The Laplace preference function trades-off accuracy against generality. It favors class descriptions that cover more positive objects over class descriptions that cover fewer, and favors class descriptions for which a lower proportion of the cover is negative over those for which it is higher. In the following study, the Laplace preference function was employed with a pruning mechanism at Step 10a of the OPUS$^o$ algorithm that pruned sections of the search space with optimistic values less than or equal to the value of a class description that covered no objects. If there was no solution with a value higher than that obtained by a class description that covered no objects, no rule was developed for the class.

The optimistic value function is derived from the observation that the cover of specializations of an expression must be subsets of the cover of that expression. Thus, specializations of an expression may not cover more positive objects, but may cover fewer negative objects than are covered by the original expression. As the Laplace preference function is maximized when positive cover is maximized and negative cover is minimized, no specialization of the expression at a node may have higher value than that obtained with the positive cover of that expression and the smallest negative cover within the sub-tree below the node. The smallest negative cover within a sub-tree below a node $n$ is obtained by the expression formed by applying all operators active at $n$ to the expression at $n$.

Other pruning can be performed through the application of $cannotImprove(n_1, n_2)$, a boolean function that is true of any two nodes $n_1$ and $n_2$ in the search tree such that $n_2$ is either the child or sibling of $n_1$ and no specialization of $n_2$ may have a higher value than the highest value in the search tree below $n_1$ inclusive but excluding the search tree below $n_2$. This function may be defined as

$$cannotImprove(x, y) \leftarrow neg(x) \subseteq neg(y) \wedge pos(x) \supseteq pos(y)$$

where $neg(n)$ denotes the set of negative objects covered by the description for node $n$ and $pos(n)$ denotes the set of positive objects covered by the description for node $n$. If $cannotImprove(n_1, n_2)$ then search below $n_2$ cannot lead to a higher valued result than can be obtained by search through specialization's of $n_1$ excluding nodes in the search space below $n_2$. This can be shown where $n_1$ is the parent and $n_2$ is the child node as follows. If $n_1$ is the parent of $n_2$ then the expression for $n_2$ must be a specialization of the expression for $n_1$ and all operators available for $n_2$ must be available for $n_1$. For any expression $g$ and its specialization, $s$, if $neg(g) \subseteq neg(s)$ then $neg(g) = neg(s)$ (as specialization can only decrease cover). It follows that for any further specialization of $n_2$, $n_3$, obtained by applications of operators $O$, there must be a specialization of $n_1$ obtained by application of operators $O$, $n_4$, which is a generalization of $n_3$ and which has identical negative cover to $n_3$. As $n_4$ is a generalization of $n_3$, it must cover all positive objects covered by $n_3$. Therefore, $n_4$ must have equal or greater positive cover and equal negative cover to $n_3$ and consequently must have an equal or greater value. It follows that it must be possible to reach from $n_1$ a node of at least as great a value as the greatest valued node below $n_2$ without applying the operator that led from $n_1$ to $n_2$.

Next we consider the case where $n_1$ and $n_2$ are siblings. It follows from the definition of $cannotImprove$ that $neg(n_1) \subseteq neg(n_2)$ and $pos(n_1) \supseteq pos(n_2)$. Let the operators $o_1$ and $o_2$ be those that led from the parent node $p$ to $n_1$ and $n_2$, respectively. It follows that $o_2$ cannot exclude any negative objects from expressions below $p$ not also excluded by $o_1$ and that $o_1$ cannot exclude any positive objects from expressions below $p$ not also excluded by $o_2$. Therefore, application of $o_2$ below $n_1$ will have no effect on the negative cover of the expression but may reduce positive cover. For any expression $e$ reached below $n_2$ by a sequence of operator applications $O$, application of $O$ to $n_1$ cannot result in an expression with lower positive or higher negative cover than that of $e$.

Table 1: Summary of experimental data sets

| Domain | Description | # Values | # Objects | # Classes |
|---|---|---|---|---|
| Audiology | Medical diagnosis | 162 | 226 | 24 |
| House Votes 84 | Predict US Senators' political affiliation from voting record. | 48 | 435 | 2 |
| Lenses | Spectacle lens prescription. | 9 | 24 | 3 |
| Lymphography | Medical diagnosis. | 60 | 148 | 4 |
| Monk 1 | Artificial data. | 17 | 556 | 2 |
| Monk 2 | Artificial data. | 17 | 601 | 2 |
| Monk 3 | Artificial data. | 17 | 554 | 2 |
| Multiplexor (F11) | Artificial data, requiring disjunctive concept description. | 22 | 500 | 2 |
| Mushroom | Identify poison mushrooms. | 126 | 8124 | 2 |
| Primary Tumor | Medical diagnosis. | 42 | 339 | 22 |
| Slovenian Breast Cancer | Medical prognosis. | 57 | 286 | 2 |
| Soybean Large | Botanical diagnosis. | 135 | 307 | 19 |
| Tic Tac Toe | Identify won or lost positions. | 27 | 958 | 2 |
| Wisconsin Breast Cancer | Medical diagnosis. | 91 | 699 | 2 |

The *cannotImprove* function was employed to prune nodes at Step 8 of the OPUS$^o$ algorithm.

## 6.2 Experimental Method

This search was performed on fourteen data sets from the UCI repository of machine learning databases (Murphy & Aha, 1993). These were all the data sets from the repository that the researcher could at the time of the experiments identify as capable of being readily expressed as a categorical attribute-value learning tasks. These fourteen data sets are described in Table 1. The number of attribute values (presented in column 3) treats missing values as distinct values. The space of class descriptions that OPUS considers for each domain (and hence the size of the search space examined for each pure conjunctive rule developed) is $2^n$, where $n$ is the number of attribute values. Thus, for the Audiology domain, for each class description developed, the search space was of size $2^{162}$. Columns 4 and 5 present the number of objects and number of classes represented in the data set, respectively.

The search was repeated once for each class in each data set. For each such search, the objects belonging to the class in question were treated as the positive objects and all other objects in the data set were treated as negative objects. This search was performed using each of the following search methods—OPUS[o]; OPUS[o] without optimistic pruning; OPUS[o] without other pruning; OPUS[o] without optimistic reordering; and fixed-order search, such as performed by Clearwater and Provost (1990), Rymon (1993), Schlimmer (1993), Segal and Etzioni (1994) and Webb (1990).

Optimistic pruning was disabled by removing the condition from Step 10a of the OPUS[o] algorithm. In other words, Step 10(a)i was always performed.

Other pruning was disabled by removing Step 8 from the OPUS[o] algorithm.

Optimistic reordering was disabled by changing Step 9 to process each node in a predetermined fixed-order, rather than in order by optimistic value. Under this treatment, the topology of the search tree is organized in a fixed-order, but operators that are pruned at a node are removed from consideration in the entire subtree below that node.

Fixed-order search was emulated by disabling Step 8b and disabling optimistic reordering, as described above.

All of the algorithms are to some extent under-specified. OPUS[o], no optimistic pruning and no other pruning are all leave unspecified the order in which operators leading to nodes with equal optimistic values should be considered at Step 9. Such ambiguities were resolved in the following experiments by ordering operators leading to nodes with higher actual values first. Where two operators tied on both optimistic and actual values, the operator mentioned first in the names file that describes the data was selected first.

No optimistic reordering and fixed-order search both leave unspecified the fixed-order that should be employed for traversing the search space. As fixed-order search is representative of previous approaches to unordered search employed in machine learning, and thus it is important to obtain a realistic evaluation of its performance, ten alternative random orders were generated and all employed for each fixed-order search task. While, due to the high variability in performance under different orderings, it would have been desirable to explore more than ten alternative orderings, this was infeasible due to the tremendous computational demands of this algorithm. The comparison with no optimistic reordering was considered less crucial, as it is used solely to evaluate the effectiveness of one aspect of the OPUS[o] algorithm, and thus, due to the tremendous computational expense of this algorithm, a single fixed ordering was used, employing the order in which attribute values are mentioned in the names file.

All of the algorithms leave unspecified the order in which nodes with equal optimistic values should be selected from $OPEN$ under best-first search, or directly expanded under depth-first search. Under best first search nodes with equal optimistic values were removed from $OPEN$ in a last-in-first-off order. Under depth-first search, nodes with equal optimistic value were expanded in the same order as was employed for allocating operators at Step 9.

Note that the fixed-order search and OPUS[o] with disabled optimistic reordering conditions both used optimistic and other pruning. Note also that while fixed-order search ordered the topology of the search tree in the manner depicted in Figure 4, it explored that tree in either a best or depth-first manner.

## 6.3 Experimental Results

Tables 2 and 3 present the number of nodes examined by each search in this experiment. For each data set the total number of nodes explored under each condition is indicated.

Table 2: Number of nodes explored under best-first search

| Data set | OPUS$^O$ | No optimistic pruning | No other pruning | No optimistic reordering | Fixed-order (mean) |
|---|---|---|---|---|---|
| Audiology | 7,044 | — | 24,199 | — | — |
| House Votes 84 | 533 | 661 | 554 | 355,040 | 1,319,911 |
| Lenses | 41 | 176 | 41 | 38 | 64 |
| Lymphography | 1,142 | 1,143 | 1,684 | 658,335 | 2,251,652 |
| Monk 1 | 357 | 9,156 | 371 | 925 | 788 |
| Monk 2 | 4,326 | 6,578 | 4,335 | 10,012 | 5,895 |
| Monk 3 | 281 | 25,775 | 281 | 682 | 656 |
| Multiplexor (F11) | 2,769 | 96,371 | 2,769 | 4,932 | 4,948 |
| Mushroom | 391 | 392 | 788 | 233,579 | — |
| Primary Tumor | 10,892 | 10,893 | 13,137 | 4,242,978 | 29,914,840 |
| Slovenian B. C. | 17,418 | 4,810,129 | 32,965 | — | 42,669,822$^\dagger$ |
| Soybean Large | 8,304 | 8338 | 21,418 | 21,551,436 | — |
| Tic Tac Toe | 2,894 | 4,222,641 | 2,902 | 16,559 | 16,471 |
| Wisconsin B. C. | 447,786 | — | 1,159,011 | — | — |

— Execution terminated after exceeding virtual memory limit of 250 megabytes.

$\dagger$ Only one of ten runs completed.

For fixed-order search, the mean of all ten runs is presented. Tables 4 and 5 present for fixed-order search the number of runs that completed successfully, the minimum number of nodes examined by a successful run, the mean number of nodes examined by successful runs (repeated from Tables 2 and 3) and the standard deviations for those runs. Every node generated at Step 7a is counted in the tally of the number of nodes explored. A hyphen (—) indicates that the search could not be completed as the number of open nodes made the system exceed a predefined virtual memory limit of 250 megabytes. An asterisk (*) indicates that the search was terminated due to exceeding a pre-specified compute time limit of twenty-four CPU hours. (For comparison, the longest CPU time taken for any data set by OPUS$^O$ was sixty-seven CPU seconds on the Wisconsin Breast Cancer data under depth-first search.)

It should be noted that one pure conjunctive rule was developed for each class. As a separate search was performed for each rule, the number of searches performed equals the number of classes. Thus, for the Audiology data using best-first search OPUS$^O$ explored just 7,044 nodes to perform 24 admissible searches of the $2^{162}$ node search space.

For only two search tasks does OPUS$^O$ with best-first search explore more nodes than an alternative. For the Lenses data, OPUS$^O$ explores 41 nodes while no optimistic reordering explores 38. For the Monk 2 data, OPUS$^O$ explores 4,326 nodes while the best of ten fixed-order runs with different random fixed orders explores 4,283. It is possible that these outcomes have arisen from situations where two sibling nodes share the same optimistic value. In such a case, if two approaches each select different nodes to expand first, one

Table 3: Number of nodes explored under depth-first search

| Data set | OPUS$^o$ | No optimistic pruning | No other pruning | No optimistic reordering | Fixed-order (mean) |
|---|---|---|---|---|---|
| Audiology | 7,011 | * | 17,191 | 3,502,475 | * |
| House Votes 84 | 568 | 17,067,302 | 596 | 10,046 | 3,674,418 |
| Lenses | 38 | 513 | 38 | 38 | 66 |
| Lymphography | 1,200 | 39,063,303 | 1,825 | 728,276 | 22,225,745 |
| Monk 1 | 364 | 54,218 | 378 | 980 | 1,348 |
| Monk 2 | 16,345 | 85,425 | 16,427 | 12,879 | 12,791 |
| Monk 3 | 289 | 63,057 | 289 | 588 | 1,236 |
| Multiplexor (F11) | 2,914 | 188,120 | 2,914 | 6,961 | 6,130 |
| Mushroom | 386 | * | 761 | 1,562,006 | 132,107,513$^\ddagger$ |
| Primary Tumor | 18,209 | 34,325,234 | 23,668 | 3,814,422 | 31,107,648 |
| Slovenian B. C. | 30,647 | 172,073,241 | 61,391 | 271,328,080 | 308,209,464 |
| Soybean Large | 9,562 | * | 23,860 | 17,138,467 | * |
| Tic Tac Toe | 3,876 | 11,496,736 | 4,010 | 93,521 | 110,664 |
| Wisconsin B. C. | 465,058 | * | 1,211,211 | * | * |

* Execution terminated after exceeding the 24 CPU hour limit.
$\ddagger$ Only three of ten runs completed.

Table 4: Number of nodes explored under best-first fixed-order search

| Data set | Runs | Minimum | Mean | sd |
|---|---|---|---|---|
| Audiology | 0 | — | — | — |
| House Votes 84 | 10 | 451,038 | 1,319,911 | 624,957 |
| Lenses | 10 | 51 | 64 | 9 |
| Lymphography | 10 | 597,842 | 2,251,652 | 1,454,583 |
| Monk 1 | 10 | 463 | 788 | 225 |
| Monk 2 | 10 | 4,283 | 5,895 | 931 |
| Monk 3 | 10 | 527 | 656 | 110 |
| Multiplexor (F11) | 10 | 4,210 | 4,948 | 364 |
| Mushroom | 0 | — | — | — |
| Primary Tumor | 10 | 10,552,129 | 29,914,840 | 12,390,146 |
| Slovenian B. C. | 1 | 42,669,822 | 42,669,822 | 0 |
| Soybean Large | 0 | — | — | — |
| Tic Tac Toe | 10 | 8,046 | 16,471 | 5,300 |
| Wisconsin B. C. | 0 | — | — | — |

— Execution terminated for all ten runs after exceeding the
virtual memory limit of 250 megabytes.

Table 5: Number of nodes explored under depth-first fixed-order search

| Data set | Runs | Minimum | Mean | sd |
|---|---|---|---|---|
| Audiology | 0 | * | * | * |
| House Votes 84 | 10 | 1,592,391 | 3,674,418 | 2,086,159 |
| Lenses | 10 | 50 | 66 | 12 |
| Lymphography | 10 | 484,694 | 22,225,745 | 27,250,834 |
| Monk 1 | 10 | 553 | 1,348 | 922 |
| Monk 2 | 10 | 9,274 | 12,791 | 2,686 |
| Monk 3 | 10 | 627 | 1,236 | 891 |
| Multiplexor (F11) | 10 | 4,467 | 6,130 | 1,164 |
| Mushroom | 3 | 105,859,320 | 132,107,513 | 22,749,211 |
| Primary Tumor | 10 | 10,458,421 | 31,107,648 | 14,907,744 |
| Slovenian B. C. | 10 | 110,101,761 | 308,209,464 | 303,800,659 |
| Soybean Large | 0 | * | * | * |
| Tic Tac Toe | 10 | 49,328 | 110,664 | 65,809 |
| Wisconsin B. C. | 0 | * | * | * |

* Execution terminated for all ten runs after exceeding the
24 CPU hour limit.

may turn out to be a better choice than the other, leading to the exploration of fewer nodes. To test the plausibility of this explanation, OPUS$^o$ was run again on the Lenses data set with Step 8 altered to ensure that where two siblings have equal optimistic value they are ordered in the same order as was employed with no optimistic reordering. This resulted in the exploration of just 36 nodes, fewer than any alternative. When OPUS$^o$ and fixed-order were run with fixed-order using the order of attribute declaration in the data file to determine operator order and OPUS$^o$ using the same order to order siblings with equal optimistic values, the numbers of nodes explored for the Monk 2 data were 4,302 for OPUS$^o$ and 8,812 for fixed-order search.

It is notable that this effect is only apparent for very small search spaces. This is significant because it suggests that there is only an effect of small magnitude resulting from a poor choice of node to expand when two nodes have equal optimistic value. This is to be expected. Consider the case where there are two nodes $n_1$ and $n_2$ with equal highest optimistic value, $v$, but $n_1$ leads to a goal whereas $n_2$ does not. If $n_2$ is expanded first, so long as no child of $n_2$ has an optimistic value greater than or equal to $v$, the next node to be expanded will be $n_1$, as $n_1$ will now be the node with the highest optimistic value. (If a child of $n_2$ has an optimistic value greater than $v$, the optimistic evaluation function cannot be very good, as the fact that $n_2$ had an optimistic value of $v$ means that no node below $n_2$ can have a value greater than $v$.) Thus, the number of unnecessary node expansions due to this effect can never exceed the number of times that nodes with equal highest optimistic values are encountered during the search.

In contrast to the case with best-first search, as discussed in Section 5.4, OPUS is only heuristic with respect to minimizing the number of nodes expanded under depth-first search.

Nonetheless, for only one search task, the Monk 2 data set, does OPUS[o] explore more nodes under depth-first search (16,345) than an alternative (both no optimistic reordering and fixed-order search that explore 12,879 and 12,791 nodes respectively). These results demonstrate that this heuristic is not optimal for this data. It should be noted, however, that the single exception for depth-first search again occurs only for a relatively small search space. This suggests that efficient exploration of the search space below a poor choice of node can do much to minimize the damage done by that poor choice, even when there is no backtracking as is the case for depth-first search.

For five data sets (House Votes 84, Lymphography, Mushroom, Primary Tumor and Soybean Large), disabling optimistic pruning has little effect under best-first search. Disabling optimistic pruning always has large effect under depth-first search. Under best-first search the smallest increase caused by disabling optimistic pruning is an increase of just one node for both the Lymphography and Mushroom data sets. Of those data sets for which it was possible to complete the search without optimistic pruning, the biggest effect was an almost 1,500 fold increase in the number of nodes explored for the Tic Tac Toe data. Under depth-first search, of those data sets for which processing could be completed without optimistic pruning, the smallest increase was a five-fold increase for the Monk 2 data and the largest increase was a 30,000 fold increase for the Lymphography data.

For seven data sets (House Votes 84, Lenses, Monk 1, Monk 2, Monk 3, F11 Multiplexor, Tic Tac Toe) disabling other pruning had little or no effect under best-first or depth-first search. The largest effects are 2.5 fold increases for the Soybean Large and Wisconsin Breast Cancer data sets under best-first search and for the Audiology, Soybean Large and Wisconsin Breast Cancer data sets under depth-first search.

From these results it is apparent that while there are some data sets for which each pruning technique has little effect (so long as the other is also employed), there are also data sets for which other pruning more than halves the amount of the search space explored and data sets for which optimistic pruning reduces the amount of the search space explored to thousandths of that which would otherwise be explored.

The effect of optimistic reordering was also highly variable. For two search tasks (best-first search for the Lenses data set and depth-first search for the Monk 2 data set) its use actually resulted in a slight increase in the number of nodes explored. This is discussed above. In many cases, however, the effect of disabling optimistic reordering was far greater than that of disabling optimistic pruning. Processing could not be completed without optimistic reordering for three of the best-first search tasks and one of the depth-first search tasks. Of those tasks for which search could be completed, the largest effect for best-first search was a 2,500 fold increase in the number of nodes explored for the Soybean Large data. Under depth-first search, of those tasks for which search could be completed, the largest effect was an 8,000 fold increase for the Slovenian Beast Cancer data. While it would be desirable to evaluate the effect of alternative fixed-orderings of operators on these results, it seems that optimistic reordering is critical to the general success of the algorithm.

For all but one data set (the Monk 2 data under depth-first search), fixed-order search on average explores substantially more nodes than OPUS[o]. It was asserted in Section 5.4 that the average case advantage from the use of OPUS[o] as opposed to fixed-order search will tend to grow exponentially as the number of search operators increases. The number of search operators for the search tasks above equal the number of attribute values in the corresponding data sets. Analysis of Tables 2 and 3 reveals that the relative advantage to OPUS$^o$ for the data sets with fewest attribute values (Lenses, Monk 1, 2 and 3, and F11 Multiplexor) is approximately a two-fold reduction in the number of nodes examined. As the number of attribute values increases, so does the relative advantage. For the four data sets with the greatest number of attribute values (Audiology, Mushroom, Soybean Large and Wisconsin Breast Cancer) in only one case (depth-first search of the Mushroom data) does the fixed-order search terminate. In this one case, OPUS$^o$ enjoys a 350,000-fold advantage. These results lend credibility to the claim that OPUS$^o$'s average case advantage over fixed-order search is exponential with respect to the size of the search space. This is illustrated in Figure 11. In this figure, for searches for which fixed-order search terminated within the resource constraints, the size of the search space is plotted against $log_2(f/o)$ where $f$ is the number of nodes explored by fixed-order search and $o$ is the number of nodes explored by OPUS$^o$.

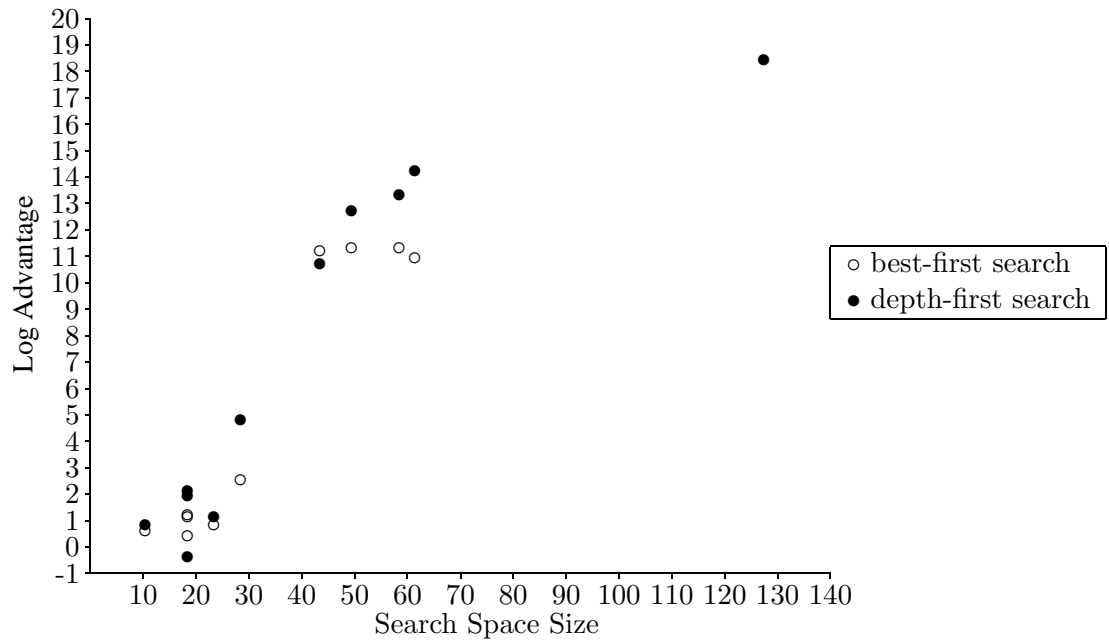

Figure 11: Plot of difference in nodes explored by fixed-order and OPUS$^o$ search against search space size.

It seems clear from these results that admissible fixed-order search is not practical for many of these search tasks within the scope of current technology.

It is interesting to observe that under best-first search, for all of the four artificial data sets (Monk 1, Monk 2, Monk 3 and F11 Multiplexor) fixed-order search often explores slightly fewer nodes than OPUS$^o$ with optimistic reordering disabled. The difference between these two types of search is that the latter deletes pruned operators from the sets of active operators under higher ordered operators whereas the former does not. Thus the latter prunes more nodes from the search tree with each pruning operation. It seems counter-intuitive that this increased pruning should sometimes lead to the exploration of more nodes. To understand this effect it is necessary to recall that other pruning can prune solutions from the search tree so long as there are alternative solutions available. For the artificial data sets in question, retaining alternative solutions in the search tree in some cases leads to a slight increase in search efficiency as the alternative can be encountered earlier than the first solution. Despite this minor advantage for a number of artificial data sets to fixed-order search over OPUS$^o$ with optimistic reordering disabled, the latter enjoys a large advantage for all other data sets for which processing could be completed. For the House Votes 84 data, fixed-order search explores over 3.5 times as many nodes under best-first search and over 350 times as many nodes under depth-first search.

It can be seen that there is some reason to believe that the the average case number of nodes explored by OPUS$^o$ is only polynomial with respect to the search space size for these machine learning search tasks. The numbers of nodes explored for the three largest search spaces are certainly not suggestive of an exponential explosion in the numbers of nodes examined (Audiology—$2^{162}$ nodes in the search space: 7,044 and 7,011 nodes examined. Soybean Large—$2^{135}$ nodes in the search space: 8,304 and 9,562 nodes examined. Mushroom—$2^{126}$ nodes in the search space: 391 and 386 nodes examined.)

It is interesting that there is little difference in the number of nodes explored by OPUS$^o$ using either best or depth-first search for most data sets. Surprisingly, slightly fewer nodes are explored by depth-first search for three of the data sets (Audiology, Lenses and Mushroom). This will be for similar reasons to those presented above in the context of the occasional slight advantage enjoyed by fixed-order search over OPUS$^o$ with optimistic reordering disabled. In some cases depth-first search fortuitously encounters alternative solutions to those found by best-first search. To evaluate the plausibility of this explanation, OPUS$^o$ was run on the three data sets in question using the fixed-order ordering to order operators with equal optimistic values. The resulting numbers of nodes explored were Audiology: 6678, Lenses: 36 and Mushroom: 385. As can be seen, these numbers are in all cases lower than the numbers of nodes explored under depth-first search. As is the case when OPUS$^o$ was outperformed by other best-first strategies, this effect appears to be of small magnitude and thus is only significant where small numbers of nodes need to be explored. For four of the data sets depth-first search explores substantially more nodes than best-first search (Slovenian Breast Cancer, 75%; Monk 2, 275%; Primary Tumor, 67%; and Tic Tac Toe, 33%).

## 6.4 Summary of Experimental Results

The experiments demonstrate that admissible search for pure conjunctive classifiers is feasible using OPUS$^o$ for the types of learning task contained in the UCI repository.

They also support the theoretical findings that OPUS$^o$ will in general explore fewer nodes than fixed-order search and that the magnitude of this advantage will tend to grow exponentially with respect to the size of the search space.

Optimistic pruning and other pruning are both demonstrated to individually provide large decreases in the number of nodes explored for some search spaces but to have little effect for others. Optimistic reordering is demonstrated to have a large impact upon the number of nodes explored.

The results with respect to the search of the largest search spaces suggest that the average case complexity of the algorithm is less than exponential with respect to search space size.

## 7. Summary and Future Research

The OPUS algorithms have potential application in many areas of endeavor. They can be used to replace admissible search algorithms for unordered search spaces that maintain explicit lists of pruned nodes, such as currently used in ATMS (de Kleer, 1986). They may also support admissible search in a number of application domains, such as learning classifiers that are inconsistent with a training set, that have previously been tackled by heuristic search.

In addition to their applications for admissible search, the OPUS algorithms may also be used for efficient non-admissible search through the application of non-admissible pruning rules. The OPUS$^o$ algorithm is also able to return a solution if prematurely terminated at any time, although this solution may be non-optimal.

The availability of admissible search is an important step forward for machine learning research. While the studies in this paper have employed OPUS$^o$ to optimize the Laplace preference function, the algorithm could be used to optimize any learning bias. This means that for the first time it is possible to isolate the effect of an explicit learning bias from any implicit learning bias that might be introduced by a heuristic search algorithm and its interaction with that explicit bias.

The application of OPUS$^o$ to provide admissible search in machine learning has already proved to be productive. Webb (1993) used OPUS$^o$ to demonstrate that heuristic search that fails to optimize the Laplace accuracy estimate within a covering algorithm frequently results in the inference of better classifiers than found by admissible search that does optimize this preference function. It was to explain this result that Quinlan and Cameron-Jones (1995) developed their theory of oversearching.

The research reported herein has demonstrated that OPUS can provide efficient admissible search for pure conjunctive classifiers on all categorical attribute-value data sets in the UCI repository. It would be interesting to see if the techniques can be extended to more powerful machine learning paradigms such as continuous attribute-value and first-order logic domains.

The research has also demonstrated the power of pruning. This issue has been given scant attention in the context of search for machine learning. Although it is presented here in the context of admissible search, the pruning rules presented are equally applicable to heuristic search. The development of these and other pruning rules may prove important as machine learning tackles ever more complex search spaces.

OPUS provides efficient admissible search in unordered search spaces. When creating a machine learning system it is necessary to consider not only what to search for (the explicit learning biases) but also how to search for it (appropriate search algorithms). It has been assumed previously that such algorithms must necessarily be heuristic techniques for approximating the desired explicit biases. Admissible search decouples these two issues by removing confounding factors that may be introduced by the search algorithm. By guaranteeing that the search uncovers the defined target, admissible search makes it possible to systematically study explicit learning biases. By supporting efficient admissible search, OPUS for the first time brings to machine learning the ability to clearly and explicitly manipulate the precise inductive bias employed in a complex machine learning task.

## Acknowledgements

This research has been supported by the Australian Research Council. I am grateful to Riichiro Mizoguchi for pointing out the potential for application of OPUS in truth maintenance. I am also grateful to Mike Cameron-Jones, Jon Patrick, Ron Rymon, Richard Segal, Jason Wells, Leslie Wells and Simon Yip for numerous helpful comments on previous drafts of this paper. I am especially indebted to my anonymous reviewers whose insightful, extensive and detailed comments greatly improved the quality of this paper.

The Breast Cancer, Lymphography and Primary Tumor data sets were provided by the Ljubljana Oncology Institute, Slovenia. Thanks to the UCI Repository, its maintainers, Patrick Murphy and David Aha, and its donors, for providing access to the data sets used herein.

## References

Buchanan, B. G., Feigenbaum, E. A., & Lederberg, J. (1971). A heuristic programming study of theory formation in science. In *IJCAI-71*, pp. 40–50.

Clark, P., & Boswell, R. (1991). Rule induction with CN2: Some recent improvements. In *Proceedings of the Fifth European Working Session on Learning*, pp. 151–163.

Clark, P., & Niblett, T. (1989). The CN2 induction algorithm. *Machine Learning*, *3*, 261–284.

Clearwater, S. H., & Provost, F. J. (1990). RL4: A tool for knowledge-based induction. In *Proceedings of Second Intl. IEEE Conf. on Tools for AI*, pp. 24–30 Los Alamitos, CA. IEEE Computer Society Pres.

de Kleer, J. (1986). An assumption-based TMS. *Artificial Intelligence*, *28*, 127–162.

de Kleer, J., Mackworth, A. K., & Reiter, R. (1990). Characterizing diagnoses. In *Proceedings AAAI-90*, pp. 324–330 Boston, MA.

Hart, P., Nilsson, N., & Raphael, B. (1968). A formal basis for the heuristic determination of minimum cost paths. *IEEE Transactions on System Sciences and Cybernetics*, *SSC-4*(2), 100–107.

Hirsh, H. (1994). Generalizing version spaces. *Artificial Intelligence*, *17*, 5–46.

Lawler, E. L., & Wood, D. E. (1966). Branch and bound methods: A survey. *Operations Research*, *149*, 699–719.

Michalski, R. S. (1984). A theory and methodology of inductive learning. In Michalski, R. S., Carbonell, J. G., & Mitchell, T. M. (Eds.), *Machine Learning: An Artificial Intelligence Approach*, pp. 83–129. Springer-Verlag, Berlin.

Mitchell, T. M. (1977). Version spaces: A candidate elimination approach to rule learning. In *Proceedings of the Fifth International Joint Conference on Artificial Intelligence*, pp. 305–310.

Mitchell, T. M. (1980). The need for biases in learning generalizations. Technical report CBM-TR-117, Rutgers University, Department of Computer Science, New Brunswick, NJ.

Moret, B. M. E., & Shapiro, H. D. (1985). On minimizing a set of tests. *SIAM Journal on Scientific and Statistical Computing*, *6*(4), 983–1003.

Murphy, P., & Aha, D. (1993). UCI repository of machine learning databases. [Machine-readable data repository]. University of California, Department of Information and Computer Science, Irvine, CA.

Murphy, P., & Pazzani, M. (1994). Exploring the decision forest: An empirical investigation of Occam's Razor in decision tree induction. *Journal of Artificial Intelligence Research*, *1*, 257–275.

Narendra, P., & Fukunaga, K. (1977). A branch and bound algorithm for feature subset selection. *IEEE Transactions on Computers*, *C-26*, 917–922.

Nilsson, N. J. (1971). *Problem-solving Methods in Artificial Intelligence*. McGraw-Hill, New York.

Oblow, E. M. (1992). Implementing Valiant's learnability theory using random sets. *Machine Learning*, *8*, 45–73.

Pearl, J. (1984). *Heuristics: Intelligent Search Strategies for Computer Problem Solving*. Addison-Wesley, Reading, Mass.

Plotkin, G. D. (1970). A note on inductive generalisation. In Meltzer, B., & Mitchie, D. (Eds.), *Machine Intelligence 5*, pp. 153–163. Edinburgh University Press, Edinburgh.

Quinlan, J. R., & Cameron-Jones, R. M. (1995). Oversearching and layered search in empirical learning. In *IJCAI'95*, pp. 1019–1024 Montreal. Morgan Kaufmann.

Reiter, R. (1987). A theory of diagnosis from first principles. *Artificial Intelligence*, *32*, 57–95.

Rymon, R. (1992). Search through systematic set enumeration. In *Proceedings KR-92*, pp. 268–275 Cambridge, MA.

Rymon, R. (1993). An SE-tree based characterization of the induction problem. In *Proceedings of the 1993 International Conference on Machine Learning* San Mateo, Ca. Morgan Kaufmann.

Schlimmer, J. C. (1993). Efficiently inducing determinations: A complete and systematic search algorithm that uses optimal pruning. In *Proceedings of the 1993 International Conference on Machine Learning*, pp. 284–290 San Mateo, Ca. Morgan Kaufmann.

Segal, R., & Etzioni, O. (1994). Learning decision lists using homogeneous rules. In *AAAI-94*.

Smyth, P., & Goodman, R. M. (1992). An information theoretic approach to rule induction from databases. *IEEE Transactions on Knowledge and Data Engineering*, *4*(2), 301–316.

Webb, G. I. (1990). Techniques for efficient empirical induction. In Barter, C. J., & Brooks, M. J. (Eds.), *AI'88 – Proceedings of the Third Australian Joint Conference on Artificial Intelligence*, pp. 225–239 Adelaide. Springer-Verlag.

Webb, G. I. (1993). Systematic search for categorical attribute-value data-driven machine learning. In Rowles, C., Liu, H., & Foo, N. (Eds.), *AI'93 – Proceedings of the Sixth Australian Joint Conference on Artificial Intelligence*, pp. 342–347 Melbourne. World Scientific.

Webb, G. I. (1994a). Generality is more significant than complexity: Toward alternatives to Occam's Razor. In Zhang, C., Debenham, J., & Lukose, D. (Eds.), *AI'94 – Proceedings of the Seventh Australian Joint Conference on Artificial Intelligence*, pp. 60–67 Armidale. World Scientific.

Webb, G. I. (1994b). Recent progress in learning decision lists by prepending inferred rules. In *SPICIS'94: Proceedings of the Second Singapore International Conference on Intelligent Systems*, pp. B280–B285 Singapore.